\documentclass[times,onecolumn,preprint,authoryear]{elsarticle}

\usepackage{prletters}
\usepackage{framed}
\usepackage{graphicx}
\usepackage{amssymb}
\usepackage{latexsym}

\usepackage{url}
\usepackage{xcolor}
\usepackage{multirow}
\usepackage{amsmath}
\definecolor{newcolor}{rgb}{.8,.349,.1}

\DeclareMathOperator*{\argmax}{arg\,max}

\journal{Pattern Recognition Letters}

\begin{document}

\clearpage

\ifpreprint
  \setcounter{page}{1}
\else
  \setcounter{page}{1}
\fi

\begin{frontmatter}

\title{A Unified Multilingual Handwriting Recognition System using multigrams sub-lexical units}

\author[1]{Wassim \snm{Swaileh}\corref{cor1}} 
\cortext[cor1]{Corresponding author: 
  }
\ead{wassim.swaileh2@univ-rouen.fr}
\author[1]{Yann \snm{Soullard}}
\author[1]{Thierry \snm{Paquet}}

\address[1]{LITIS Laboratory - EA 4108 
Normandie University - University of Rouen 
Rouen, France}

\received{19 November 2017}
\finalform{-}
\accepted{-}
\availableonline{-}
\communicated{-}

\begin{abstract}
We address the design of a unified multilingual system for handwriting recognition. Most of multilingual systems rests on specialized models that are trained on a single language and one of them is selected at test time. While some recognition systems are based on a unified optical model, dealing with a unified language model remains a major issue, as traditional language models are generally trained on corpora composed of large word lexicons per language. Here, we bring a solution by considering language models based on sub-lexical units, called multigrams. Dealing with multigrams strongly reduces the lexicon size and thus decreases the language model complexity. This makes possible the design of an end-to-end unified multilingual recognition system where both a single optical model and a single language model are trained on all the languages. We discuss the impact of the language unification on each model and show that our system reaches state-of-the-art methods performance with a strong reduction of the complexity. 

\end{abstract}

\begin{keyword}
Sub-lexical units \sep Multilingual \sep Language model \sep Handwriting recognition  \sep Multigrams
\end{keyword}

\end{frontmatter}

\section{Introduction}
\label{sec:Intro}






Offline handwriting recognition is a challenging task due to the high variability of data, as writing styles, the quality of the input image and the lack of contextual information. The recognition is commonly done in two steps. First, an optical character recognition (OCR) model is optimized for recognizing sequences of characters from input images of handwritten texts~(\cite{plotz2009markov,singh2013optical}). Second, a language model is used to model language constraints~(\cite{mikolov2010recurrent,croft2013language, mousa2014sub}). At decoding step, both models are combined to get a prediction. 
Recent improvements in deep learning techniques~(\cite{graves2013speech,lecun2015deep,schmidhuber2015deep}) have enhanced the capacity of optical models. In contrast, only few works have been dedicated to language models for handwriting recognition. They are generally trained to model sequences of words and sequences or characters (\cite{messina2014over}). 

Most recognition systems are defined to recognize texts written in a specific language and only few works proposed multilingual recognition systems. A multilingual recognition system allows to process documents written in various languages, without prior knowledge about the language. They are of two types. The first one consists in defining one specific system per language and selecting one of them to get a prediction (\cite{miguel2012character,mathew2016multilingual}). The second one consists in defining a unified recognition system where at least a part of the system is trained on several languages. Recent works proposed unified OCR models, by merging the character sets of every languages, as in  \cite{bluche2017gated,moysset2014a2ia,kozielski2014open}. Some characters are often shared by languages of the same origin, which may be beneficial to improve recognition performance. 
For instance, Latin languages share at least $21$ characters~(\cite{diringer1951alphabet}) while $14$ Arabic and Persian languages share at least $28$ characters (\cite{margner2012guide}). 

Designing a unified language model is less straightforward than designing a unified optical model. This is due to the word lexicons sizes which are often large (to model as much as possible each language). Thus, combining word lexicons from various languages strongly increases the model complexity and may become intractable. One solution is to consider a lexicon of characters but language models based on characters perform often poorly (\cite{plotz2009markov}). 

Designing a unified language model is attractive for taking into account similarities between languages of the same origin that exist at the lexical, morphological or syntactic levels (\cite{kalindra2004some}). In this respect, language models based on sub-lexical units have been recently proposed for handwritten recognition (\cite{swaileh2017multigram}, \cite{swaileh2016unified}). Dealing with sub-lexical units has a number of advantages: on the one hand, it allows to significantly reduce the lexicon size and, on the other hand, to improve the recognition rate, as it partly tackles the out-of-vocabulary words problem.

In this paper, we proposes an end-to-end unified multilingual recognition system for handwriting recognition, where both the optical model and the language model combine various languages. Both models rest on state-of-the-art methods. On the one hand, we make use of an optical model based on deep recurrent neural networks. On the other hand, our language model uses sequences of sub-lexical units, called multigrams. Dealing with multigrams allows to reduce the lexicon size from each language and thus to build a unified language model of reasonable size. 
We evaluate our approach on languages of the same origin (French and English), as the similarities between the languages may impact the capacity of the system. In our experiments, we show that a) combining languages of the same origin in a unified framework allows to strengthen the capacity of the optical model; b) combining the languages  does not significantly affect the robustness of the language model. This allows to build a language model estimated on all the training corpus, without the need to separate the languages; c) build a unified multilingual system seems better than dealing with specific-language systems, where one of them must be selected to provide a prediction. 

The rest of the paper is organized as follows: Section \ref{sec:rw} introduces the related works on multilingual handwriting recognition systems; In section \ref{sec:proposed_system}, we present the framework of the unified multilingual handwriting recognition system that we propose here; Then, we show and discuss experimental results where the English and French languages are combined (section \ref{sec:eval}) before concluding.

\section{Related works}
\label{sec:rw}


Multilingual handwriting recognition systems have some similarities with methods proposed for multilingual speech recognition systems, as in \cite{ghoshal2013multilingual} and \cite{kumar2005multilingual}. In general the problem can be solved using two types of approaches: the selective approaches and the unified approaches. In the following we only focus on works dedicated to handwriting recognition but the analysis stands for speech recognition as well. 


\subsection{Selective approaches}
\label{sec:selectapp}
The selective approaches are based on specialized recognition systems, i.e. each system is dedicated to one language, and the output of one of them is selected at the time of processing one specific sample (\cite{peng2013multilingual}). There are two kinds of selective approaches. The first one consists in applying specialized recognition systems in parallel to get several transcripts. The transcript with the highest confidence score is then selected. A major issue is to compare the confident scores given by each specialized system, as they span over different scales (\cite{lee1997unified}). 

The second approach is based on a language detection module, as in \cite{mathew2016multilingual} and \cite{barlas2016language}. This module aims at detecting the language of the input script. This allows to select the specialized system corresponding to the language detected. For instance, \cite{miguel2012character} applied first a probabilistic model for language identification and then a specialized language model based on characters to obtain a prediction. However, these approaches are prone to wrong language detections, which has a direct impact on the system performance (\cite{moysset2014a2ia,kozielski2014multilingual}). 

\subsection{Unified approaches}
\label{sec:unifiedapp}


Regarding the unified approaches, there is one or more components that model multiple languages in a unified framework. Three types of unified recognition systems have been presented in the literature: there are systems where 1) only the optical model is trained on multiple languages, i.e. in a unified manner, and then specialized language models are used; 2) the optical model is partially-unified as an encoding part is done on multiple languages and then decoding parts are defined for each language; 3) both the optical model and the language model are trained in a unified framework. 

\subsubsection{Systems with a unified optical model only}

Most  multilingual handwriting recognition systems are based on a unified optical model and on a specialized language model. For instance, \cite{kessentini2008multi} defined a multilingual system based on a unified Hidden Markov Model (HMM) for Arabic and Latin handwriting recognition. 
Similarly, \cite{lee1997unified} proposed an interconnected network based on Hidden Markov Models for Hangul and English languages. 

\cite{kozielski2014multilingual} presented a comparative study about the use of traditional HMMs and LSTM neural networks to construct a unified optical model for English and French. Specialized language models are used as the authors argue that the languages have different vocabularies (i.e. words). 
They evaluated the two systems on the Maurdor dataset and showed that the unified LSTM-based system outperforms the HMM-based system. However, any of the two approaches highlights the feasibility of unifying optical models of multiple languages of the same origin (Latin scripts).
Similarly, \cite{moysset2014a2ia} built a unified multilingual system for both printed and handwritten recognition. The system is composed of a unified optical model based on a Recurrent Neural Network (RNN) and of a set of specialized language models (for the English, French and Arabic languages). 


\subsubsection{Systems based on a partially-unified optical model only}
\cite{bluche2017gated} proposed a unified multilingual system where an encoder part based on gated convolutional neural network is trained in a unified manner. Then, specialized Bi-directional Long Short Memory RNNs (BLSTM-RNN) are defined for the decoding step of the optical model. Finally, the authors used specialized hybrid words/characters language models to get the final character string predictions. 

\subsubsection{Systems where both the optical and language models are unified}

\cite{ray2015ocr} introduced a bilingual recognition system consisting of a unified BLSTM optical model and a unified 3-gram language model. The training corpus contains documents written in two languages where each paragraph is related to a specific language. Four unified optical models are trained on various data representations (binarizations, segmentations), which allows to get 4 recognition hypotheses from a single test sequence. Then, a sliding window process is applied to match every 3 consecutive words from a recognition hypothesis with 3-grams estimated by the language model. A cumulative score computed on the entire hypothesis sequence allows to select the most probable hypothesis. 
This process can be seen as a verification algorithm, which limits the recognition only to the words belonging to the training corpus.

In a previous work (\cite{swaileh2016unified}), we introduced a unified bilingual recognition system based on syllables. The proposed system is composed of a unified optical model based on HMMs and a bilingual n-gram language model of syllables. In this work, French and English linguistic ressources were used to decompose words into syllables. Besides, no Out-Of-Vocabulary (OOV) words were considered by the language models. This simplifies the recognition task but this is a serious limitation. This preliminary work showed that a unified language model based on sub-lexical units (syllables) is feasible and can be envisaged as an alternative to word-based language models. A major issue concerns the linguistic expertise which is required to get word decompositions into syllables, and syllables databases are only available for a few number of languages. 


This review shows that very few works have proposed multilingual handwriting recognition systems that are completely unified. Such approaches should have the advantage to train a unique system whatever the languages that are considered. This should avoid to face the issues highlighted in the selective approaches, when the system is composed of specialized models. 

\section{Design of a unified multilingual system}
\label{sec:proposed_system}

\subsection{Unified recognition system proposal}
\label{sec:unifiedRec}

\begin{figure*}[!]
\centering
\scriptsize
\centering
\includegraphics[width=0.8\textwidth]{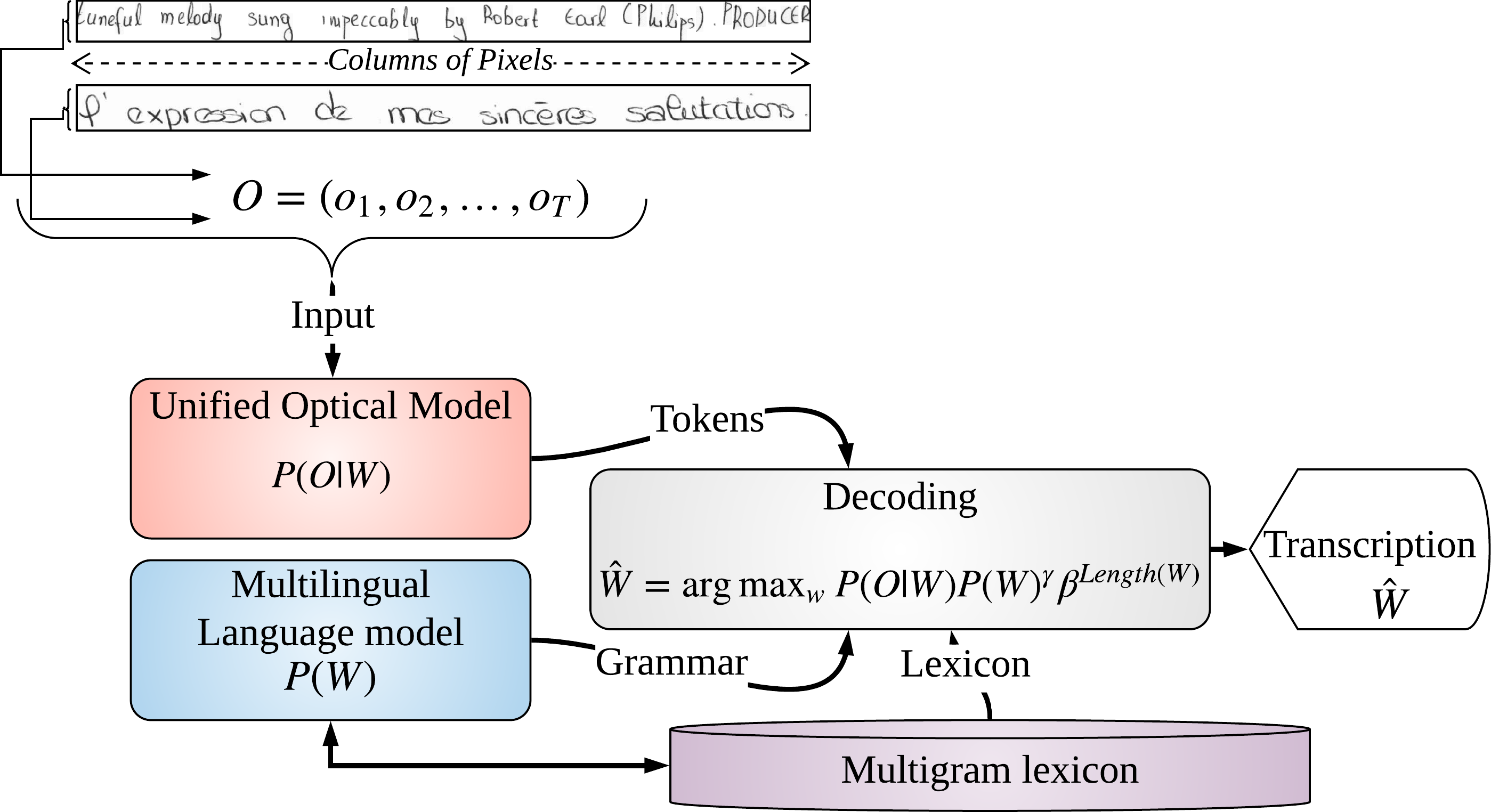}
\caption{Our unified multilingual recognition system composed of a unified optical model dealing with text images of any language and a multilingual language model based on a unified multigram lexicon. At test time, the decoding process consists in combining both models to get a transcription.}
\label{fig:sys}
\end{figure*}

Similarly to standard handwriting recognition system, our proposed system is composed of two components:~1)~an optical model which is trained to recognize a sequence of characters from a sequence of observations;~2)~a language model which is trained to model language constraints in sequences of words (Figure \ref{fig:sys}). 

Designing a unified optical model can be done easily by combining the character sets. Combining the character sets may have many advantages when the languages are of the same origin, i.e. where some characters are shared between the languages. On the one hand, the system complexity is reduced as the shared characters are modeled only once. In addition, the number of occurrences of the shared characters is larger in the training set and the optical model can take benefit from this. Moreover, we also expect the optical model to generalize better as the shared characters come from various languages, so they appear in different contexts in the text. 

In contrast to optical models, designing a unified language model is less straightforward. This is especially due to the model complexity that increases. Indeed, if one expects the multilingual language model to perform equally well for every language, then the size of the working lexicon is the sum of each lexicon size. This explains why multilingual unified systems are generally based on a unified optical model only and they use several specific language models, as it is the case in \cite{kozielski2014multilingual}, \cite{moysset2014a2ia} and \cite{bluche2017gated}. 

In this paper, we propose a unified language model based on sub-lexical units, called multigrams. Here, we consider multigrams as sequences of characters of variable length. Some examples of word decomposition into multigrams are illustrated in Table \ref{tbl:k-multigram} and the proposed system architecture is illustrated in Figure \ref{fig:sys}. Working with a language model based on multigrams has many advantages compared to language model based on words. First, the lexicon size is strongly reduced, as we only consider sub-parts of words i.e. short sequences of characters. This allows to build a unified language model based on a lexicon of reasonable size. Second, sub-lexical units can generate an open-vocabulary (i.e. words that do not belong to the training lexicon can be generated during decoding if the transitions between their multigrams have non zero probabilities). This allows to reduce the Out-Of-Vocabulary words rate and increases the effective coverage rate. Third, compared to other sub-lexical units such as syllables, multigrams do not obey any linguistic rule but statistical rules. Therefore defining the decomposition of a word into multigrams is a statistical data-driven process which does not require any linguistic expertise, and can be apply to any language. Finally, multigrams can be seen as a good trade-off between words and characters to improve the capacity of language models on specific languages (\cite{swaileh2017multigram}). The many advantages of language models based on multigrams led us to design a multilingual language model based on a unified lexicon of multigrams. To our knowledge, this is the first work that proposes an end-to-end unified multilingual system where the language model is based on sub-lexical units that can be obtained using a data-driven process for any language and which is applied in an open-vocabulary recognition task. 


\subsection{Recognition system architecture}
\label{sec:archi}

We present the various components which compose our multilingual recognition system. 

\subsubsection{Optical model}
\label{sec:ocr}
The recognition of a number of different language scripts within a unified optical model rests on three unification steps. 
The first step is to use the same observation descriptors for all script images, thus obtaining a unified feature vector. 
The second step is to define a unified character set by combining all the characters of the languages of interest. In the case of languages of the same origin, the unified character set contains fewer characters than the sum of the character sets of the different languages, owing to the shared characters. 
The third step is to train a statistical model, that relates to the unified optical model of the recognition system, using the unified character set and the unified descriptor set (feature vectors).
The training algorithm (initial parameters, estimation scheme, and training strategy) is the same for any example in the multilingual training corpus. 
Here, we use a BLSTM-RNN network trained using the Connectionist Temporal Classification approach (\cite{graves2005bidirectional, graves2006connectionist}). Details on our implementation are given in section \ref{sec:protocol}.



\subsubsection{Language model}
\label{sec:lm}
Similar to the traditional n-gram language model of words, the n-gram language model of sub-lexical units (made of multigrams) is estimated with back-off coefficients. The idea is to replace words in the traditional language model by their corresponding multigrams sub-lexical units.

Training a multigram-based n-gram language model involves three main steps (as for training any n-gram statistical model): the training corpus tokenization; the lexicon size and type determination; and the language model parameter estimation with a fixed n-gram order.
The tokenisation step replaces any word of the training corpus by its corresponding $k$-multigram decomposition, where $k$ is the multigram order (see section \ref{sec:unifiedRec}).
Table \ref{tbl:k-multigram} gives an example of word decomposition into $k$-multigrams, for various values of k.
The training corpus that contains text from two or more languages can be considered as a multilingual language model training corpus.

\begin{table}[h]
\centering
\caption{Illustration of a French (FR) and English (EN) word decomposition into $k$-multigrams with $2 \leqslant k \leqslant 5$}
\begin{tabular}{c|c|c|c|c|c|}
\cline{2-6}
\multicolumn{1}{l|}{}    & words              & 2-multigram           & 3-multigram          & 4-multigram         & 5-multigram         \\ \hline
\multicolumn{1}{|c|}{FR} & Merci   & Me r ci    & Mer ci    & Merc i    & Merci    \\ \hline
\multicolumn{1}{|c|}{EN} & darling & d ar li ng & dar l ing & dar ling & 
dar ling \\ \hline
\end{tabular}%

\label{tbl:k-multigram}
\end{table}

The second step is to fix the size of the language model lexicon.
Due to the compact size of the lexicon of $k$-multigrams, all words in the training corpus can be considered.
The multigram order $k$ defines the type of k-multigram lexicon. 

The third step is to define the order $n$ of the n-gram language model.
Then, the model parameters can be estimated using the defined lexicon and the tokenized training corpus. 
High values of n are required if one wants to ensure modeling long term dependencies.

\subsubsection{Decoding}
\label{sec:decode}
During the recognition process, the outputs of the unified optical and language models are combined into a search graph. The search graph can be decoded using dynamic programming such as the well-known Viterbi algorithm. Here the search graph is a weighted finite State Transducer (WFST). It is build by the composition of three sub-transducers representing the tokens (characters), the lexicon and the language model (grammar) respectively. The token transducer (T) represents all possible characters that can be produced by the unified optical model, from the input frames. The lexicon transducer (L) represents the possible $k$-multigrams of the languages of interest that can be produced from the character set. The language model transducer (G) represents the n-gram language model of $k$-multigrams which is trained from the multilingual training text dataset. 

According to equation \ref{eq:fst}, the composition of the tokens, lexicon and language model transducers is performed after successive application of the minimization and determination processes on the lexicon and language model transducers:
\begin{equation}
S=T \circ min(det(L \circ G ))
\label{eq:fst}
\end{equation}
where $\circ$, $det$ and $min$ denote composition, determinization and minimization respectively and $S$ denote the combined transducer. Both operators aim at reducing the branching factor of the search graph.




Decoding consists in applying the Viterbi algorithm on the combined transducer, so as to find the best (or the n best) sequence(s) of multigrams corresponding to the observation (the input observation feature sequence). Two hyper-parameters are used to guide the decoding process: the language model scale parameter $\gamma$ and the word insertion penalty parameter $\beta$ that controls the insertion of frequent short words. These two parameters need to be optimized to find an optimum coupling of the optical model with the multigram language model, because these two models are estimated independently from each other during training.

During decoding, we seek for the sentence $\hat{W}$ that maximizes the posterior probability $P(W|O)$ among all possible sentences $W$.
Using Bayes' formula and introducing the two hyper-parameters mentioned above, we obtain the formula given in equation \ref{eq:op} which governs the decoding process: 

\begin{equation}
\hat{W}=\argmax_w P(W|O)=\argmax_w P(O|W) P(W)^\gamma \beta^{Length(W)}
\label{eq:op}
\end{equation}
where $O$ is the observation sequence extracted from the image and $P(O|W)$ is the probability of the observation sequence given the sentence $W$.
$P(W)$ is the prior probability of the sentence deduced from the language model.

\subsection{Deriving multigram sub-lexical units}
\label{sec:deriveMultigrams}



In this paper, we propose a unified language model based on sub-lexical units called multigrams. We consider multigrams as sequences of characters of variable length. The interest of using such decomposition have been highlighted in subsection \ref{sec:unifiedRec}. Here, we focus on the description of the data-driven statistical model that allows to learn a set of sub-lexical units from textual data. Textual data can be either the transcriptions from the training corpus, or any other textual resource of the languages considered. The aim is to obtain a set of sub-lexical units that will serve as the main components of the multi-lingual language model in our recognition system. The problem must be considered as unsupervised, because the sequence of multigrams corresponding to a given observation sequence of character (a word or a sentence) is unknown. To address this problem, we define a generative Hidden Semi-Markovian Model (HSMM, \cite{murphy2002hidden,yu2010hidden}) where the k hidden states of the model account for the character length of the multigrams. In other words, the HSMM is trained to model the segmentation of a sequence of characters into multigrams in an unsupervised manner. As in \cite{deligne1995language}, it is assumed that the multigrams are independent from one another. This means that multigrams follow a zero-order Markov process. We have a $0/1$ duration probability of the hidden states, and there is a bijective relation between states and durations.

Let $O_{1:T} = O_{1} \ldots O_{t} \ldots O_{T}$ be an observation sequence of length $T$ ($T$ characters), where $O_t$ is the observation (character) at time $t$. One denotes $\mathcal{S}~=~\{s_1,\ldots,s_M\}$ the set of possible states and $Q_{1:T}$ a state sequence of length $T$, where each term $Q_{[t_1,t_2]}$ is an element of $\mathcal{S}$ starting at time $t_1$ and ending at time $t_2$ with a duration $d~=~t_2-t_1+1$. 
The joint probability of an observation sequence and a state sequence is expressed as follows:
\begin{align}
P(O,Q) &=\prod_{l} P({ O_{[t-d_{l}+1,t]} \; \vert \; Q_{[t-d_{l}+1,t]} }) P({Q_{[t-d_{l}+1,t]} }) \nonumber \\
	&= \prod_{l} P({ O_{[t-d_{l}+1,t]} \vert d_{l} }) \label{eq:jointProbMultigram}
\end{align}

Equation \ref{eq:jointProbMultigram} presents a way to define a Discrete Hidden Semi-Markov Model of zero order to model multigrams of variable length. The HSMM is trained using the well-known Baum-Welch algorithm, based on the Forward-Backward algorithm. 
Given a maximal length $d_{max}$ for multigrams, any possible sequence of characters for which the length does not exceed $d_{max}$ is considered in the model.


Once the HSMM has been trained, the decoding step consists in assigning a sequence of multigrams to any sequence of characters of the training corpus. In other words, the goal is to find the most probable sequence of multigrams $Q^{*}$, given any observation sequence $O_{1:T}$ of the training corpus, which is defined as follow:
\begin{equation}
Q^{*} = \underset{q_{1:T}}{\argmax} \; \Big( \; P(O_{1:T},Q_{1:T}=q_{1:T}) \Big)
\label{eq:BestPath}
\end{equation}

Equation \ref{eq:BestPath} can be solved using the Viterbi algorithm. The optimal path is obtained using a backward pass defined as follow:
\begin{align}
d_i^* &= \underset{d}{\argmax} \; \Big( \underset{d'}{\argmax} \; \big( \; \delta_{t-d}(d') \; b(O_{[t-d+1:t]})^{1/d} \big) \Big) \\
t_i^* &= t_{i-1}^* - d_i^* 
  \end{align}
where $\delta_t(d) \stackrel{\Delta}{=} \underset{q_{1:t-d}}{\max} \; P(O_{1:t},Q_{[t-d+1,t]}=d, q_{1:t-d})$ and $t_0^*=T$. The exponent term $1/d$ is a penalty term defined to favor longer multigrams at the expense of shorter ones. Indeed, the Viterbi algorithm tends to produce short multigrams because the emission term $b(O_{[t-d+1:t]})$ is related to the occurrences of multigrams and the shorter multigrams are often more frequent than the longer ones. 


We train one HSMM per language. Each one is optimized to produce the most frequent multigrams which appear in the training texts of a given language. Finally, language specific multigram lexicons are combined. After the tokenization step of the whole training corpus into multigrams, the unified language model is trained as described in section \ref{sec:lm}. At test time, the unified language model is applied on a text whatever its language, using the decoding step described in section \ref{sec:decode}.

\section{Multilingual system evaluation: English and French case study}
\label{sec:eval}

\subsection{Datasets and lexicon units properties}
\label{sec:data}


We apply our multilingual system on the English IAM (EN) and the French RIMES (FR) handwriting databases.
The RIMES database (\cite{gorski1999a2ia}) contains handwritten mails of different writers.
The IAM database (\cite{marti2002iam}) is inspired from the LOB corpus (\cite{johansson1980lob}) and it is composed of texts from different writers. 
These databases are divided into three parts: the training, the validation and the test datasets. Our unified system is trained on the combination of the FR and EN training datasets. 
The hyper-parameters $\gamma$ and $\beta$ used for decoding (see section \ref{sec:decode}) are optimized using the FR+EN validation dataset. The system is evaluated on the FR and EN specialized test datasets independently. 
Table \ref{tbl:DB} shows the number of text line images per dataset. 

\begin{table}[h]
\centering
\caption{Numbers of text line images in the FR and EN datasets used for training, validation and test.}
\begin{tabular}{|c|c|c|c|}
\hline
\multirow{2}{*}{Databases} & \multicolumn{3}{c|}{\begin{tabular}[c]{@{}c@{}}Number of text line images\\ per dataset\end{tabular}} \\ \cline{2-4} 
 & Training & Validation & Test \\ \hline
FR & 9947 & 1333 & 778 \\ \hline
EN & 6482 & 976 & 2915 \\ \hline
FR+EN & 16429 & 2309 & ---- \\ \hline
\end{tabular}

\label{tbl:DB}
\end{table}

The unified optical model is trained on the joint FR+EN training dataset. The FR dataset is composed of $100$ character classes while the EN dataset contains only $79$ character classes. $77$ of them (including white space) are shared between the two datasets so that the unified dataset sets contains $102$ character classes only. The entire EN character set is included in the FR training dataset, except the two characters "\#" and "\&". Most of the additional character classes in the FR dataset are accented characters. 

For training the multilingual language model, we used the combined training sets from the French and English languages. By nature, this dataset brings a relatively small vocabulary which we refer to as "small FR+EN". In addition, we also trained the language model on a very large dataset of text samples made of both English and French texts which we later refer to as "large FR+EN". This large dataset was collected as follows. Regarding the FR dataset, we collected $52,930$ paragraphs from French Wikipedia pages. This Wikipedia and the RIMES training datasets are joint and the $29.1$k most frequent words are used for training the language model. Regarding the EN dataset, we combined the LOB (excluding IAM validation and test examples), Brown and Wellington datasets to form the large EN lexicon used for training the language model. 

\begin{figure}[h]
\centering
\scriptsize
\centering
\includegraphics[width=0.5\columnwidth]{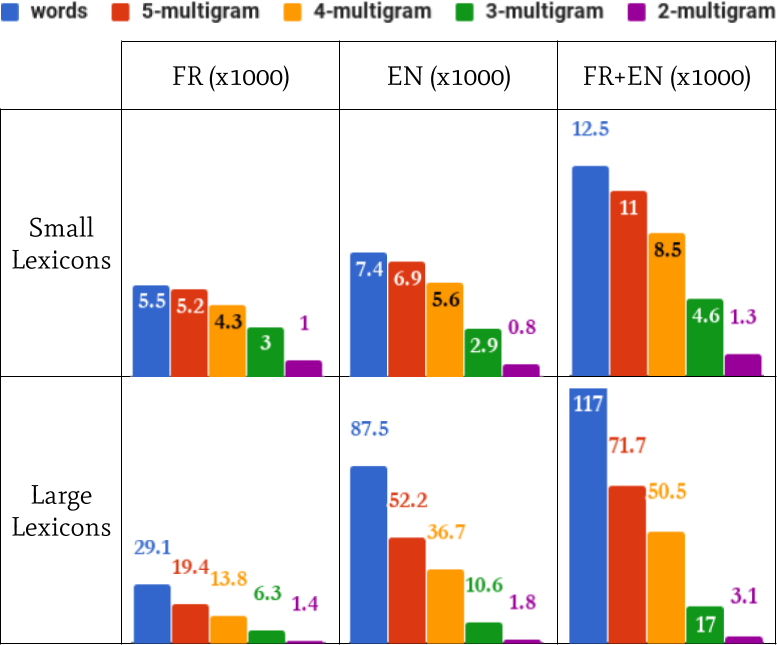}
\caption{Sizes of the small and large lexicons according to the lexical (words) or sub-lexical (multigrams) decomposition, on the French (FR), English (EN) and unified (FR+EN) training sets.}
\label{fig:Lexicon}
\end{figure}

Figure \ref{fig:Lexicon} gives the lexicon sizes when considering lexicons of k-multigrams or words. For example the small FR+EN lexicon of words contains $12.5$k words, while the corresponding 2-multigram lexicon is made of $1.3$k elements only. In contrast the multilingual "large FR+EN" lexicon contains $117k$ elements while there is only $3.1k$ 2-multigrams to describe the same dataset. This highlights a significant reduction of the lexicon size when using k-multigrams sub-lexical units instead of words.

\subsection{System configurations and Evaluation protocol}
\label{sec:protocol}

The observations given in input of our system are gray-scales images of text-lines. The input images are normalized to a fixed height of $100$ pixels, preserving the aspect ratio. The gray scale is normalized to zero mean and unity variance (standardization). 
The unified optical model is a Recurrent Neural Network (RNN) composed of 4 Bi-directional Long Short Term Memory (BLSTM) layers with $200$ cells in each direction. Our network is trained using the Connectionist Temporal Classification (CTC) approach (\cite{graves2006connectionist}). 
The model is implemented using the EESEN toolkit (\cite{miao2015eesen}) that has been first introduced for speech recognition. We used the curriculum learning algorithm for training the observation sequences from the shorter sequence to the longer one. Training starts with a learning rate of $10^{-5}$ which decreases during training. Training ends after $108$ training epochs using an early stopping criterion. 

\begin{figure*}[h!]
\centering
\scriptsize
\centering
\includegraphics[width=0.49\textwidth]{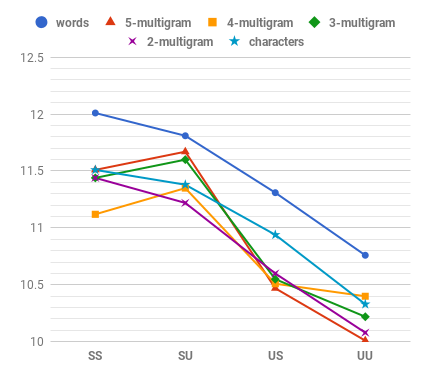}
\rule{1 pt}{200 pt} \includegraphics[width=0.49\textwidth]{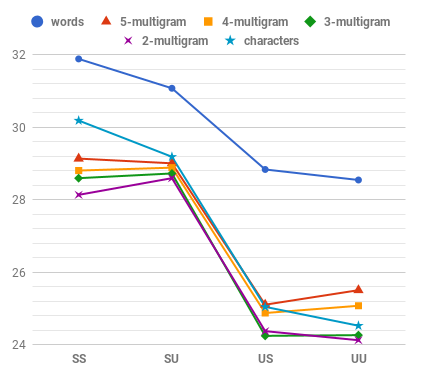}
\caption{WER (\%) on the FR (left) and EN (right) test datasets for the four experimental setups where language models are based on small vocabularies.}
\label{fig:smallFREN}
\end{figure*} 



We used 9-gram language models of $k$-multigrams or words, and 10-gram language models of characters in any of our experiments. We explored $k$-multigrams models for $2\leq k \leq 5$. To ensure that long distance dependencies are estimated in the model, all text lines are concatenated in order to form a single text line. 
Every language model is estimated using the MIT language modeling toolkit from \cite{hsu2009language}, where the back-off coefficients are estimated using the modified Kneser-Ney smoothing method (\cite{kneser1995improved}). 

Viterbi's two-pass decoding algorithm is applied at the paragraph level to track contextual dependencies as much as possible. The beam search parameter has been fixed to 0.8 in all the experimentations while language model scale and word insertion penalty are optimized for each scenario (see below). 



During test, we are interested in quantifying the contribution of each stage (i.e. optical model and language model) to the performance of the unified system. In this respect, we conducted four experiments (scenarios). 
The first one (\textbf{UU}) consists in having a whole unified system (both the optical model and the language model are unified). The second experiment (\textbf{US}) consists in combining a unified optical model with a specialized language model. The third experiment (\textbf{SU}) combines a specialized optical model with a unified language model. Finally, the last one (\textbf{SS}) is composed of specialized models (both the optical model and the language model). In the following section we report and analyze the performance of these four configuration of the system. 

\subsection{Recognition results}
\label{sec:results}

We now evaluate our approach on the small and large EN and FR vocabularies. We first present observations we obtained for the four experiments described above in sections \ref{sec:eval_small} and \ref{sec:eval_large} and then discussed the results in section \ref{sec:eval_discuss}. Finally, we compare our system with state-of-the-art methods. 


\subsubsection{Evaluation with small EN and FR lexicons}
\label{sec:eval_small}

Figure \ref{fig:smallFREN} presents the Word Error Rate (WER) obtained after training the language model on the small FR and EN datasets respectively. These figures present the performance of the systems for 6 differents types of language models: words, characters, and $k$-multigrams (with $2\leq k \leq 5$). 

One can first notice that the unified optical model (US and UU) significantly improves the specialized framework whatever the test set (FR and EN) and the configuration of the system (specialized or unified). The optical model takes advantage of language similarities to be more robust and more efficient. 

Besides, multigram-based language models often outperform traditional language models based on words or characters. Especially, the 2-multigrams language models are always better than the traditional ones. This confirms our hypothesis that multigrams are a good trade-off between words and characters for a language modeling task. Compared to the specialized frameworks, the unified scheme generally provides similar results except for the FR dataset where a slight improvement is observed when the optical model is unified. 

Finally one can notice that the performance obtained on the IAM dataset is rather low for any of the language model type (word, character or $k$-multigrams), and system unification type (SS, SU, US, UU), when a small vocabulary is used for training the language models. This is a particular difficulty encountered on the IAM dataset because of the low lexicon coverage rate of the training dataset lexicon on the test dataset. This is the reason why most studies have come to use additional linguistic resources to get better performance. To the next, we analyze tour systems using a large vocabulary setting.

\subsubsection{Evaluation with large EN and FR lexicons}
\label{sec:eval_large}

\begin{figure*}[h!]
\centering
\scriptsize
\centering
\includegraphics[width=0.49\textwidth]{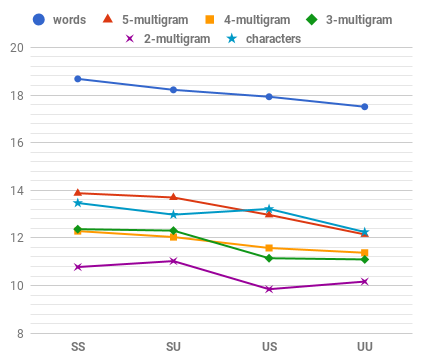}
\rule{1 pt}{200 pt} \includegraphics[width=0.49\textwidth]{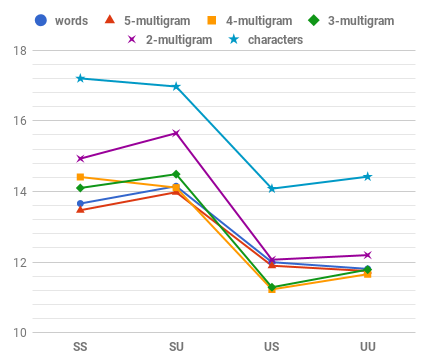}
\caption{WER (\%) on the FR (left) and EN (right) test datasets for the four experimental setups where language models are based on large vocabularies.}
\label{fig:largeFREN}
\end{figure*}

We now report experimentation results on the large EN and FR datasets described in section \ref{sec:data}. Figure \ref{fig:largeFREN} presents the Word Error Rate (WER) obtained on the FR and EN test datasets respectively. Similarly to small lexicons, we observe that unifying the optical models (US and UU) improves the recognition rate. Unifying the language models does not impact the recognition performance, even if the unified lexicons are highly imbalanced (the unified word lexicon has $116k$ words with only $29k$ from the French language). 
Moreover, dealing with multigrams language models generally reduces the WER, compared to systems based on traditional language models of words or characters. 

\subsection{Discussion}
\label{sec:eval_discuss}






We have seen that unifying English and French optical models allows to significantly improve the recognition performance of the system. This may be explained by two facts: on the one hand, there is a lot of shared characters between the two languages, which increases the number of training samples for theses character classes; on the other hand, the shared characters appear in a different context (i.e. into different words), which gives more variability in the data as well. The model can benefit of it for increasing its generalization capacities. The unified optical model achieves a Character Error Rate (CER) of $10.0\%$ alone (i.e. without language model) on the FR test set, while the specialized model has a CER of $12.3\%$. Similarly, we obtain a CER of $15.3\%$ on the EN test set with the unified optical model while the CER of the specialized model is $18.9\%$. 

Compared to the specialized language model, the contribution of a unified language model appears rather moderate, as results are generally equivalent whatever the lexicon size. In fact, there is no reason to believe that some improvements should be gain by combining the English and French languages. Indeed, the language modeling task should become more complex due to some language similarities. The key point is that unifying both languages into the language model does not affect the system performance, which allows to design a system with only one language model whatever the languages. 

There is only one case where unifying the language model improves the performance of the specialized system. This is for the UU setting on the FR test dataset with a small lexicon. In this case, the system benefits from both a better CER with a unified optical model and a better coverage of the unified language model. This improvement may be explained by the fact that the OOV rate, computed at a word-level, is reduced in the unified cases (compared to the specialized ones) and that the effective coverage rate, computed at a sub-lexical-level, is high. For instance, the OOV rate on the FR set for the 3-multigrams is of $1.4\%$ in the specialized case and of $0.7\%$ in the unified one, while the effective coverage rate for the unified dataset is equal to $98.5\%$. In contrast, the OOV rates on the EN set are very small ($0.06\%$ and $0.03\%$ for the specialized and unified case respectively) and the effective coverage rate stay low enough even if the lexicon has been unified ($89.9\%$). Besides, when lexicons are large, unifying the languages does not significantly reduce the OOV rate, which does not impact the recognition performance. For example, the unification of languages for the 3-multigrams only reduce the OOV rate from $0.5\%$ to $0.4\%$ on the FR dataset and from $2.2\%$ to $1.6\%$ on the EN dataset. 

Moreover, languages models based on multigrams often outperform traditional language models based on words or characters and there is always a system based on multigrams which is better than systems based on traditional models. Another advantage of dealing with multigrams is related to the lexicon size, which is highly reduced compared to word lexicons. For instance, there is $116k$ words in the large unified EN~+~FR~lexicon while there is $71.6k$ 5-multigrams, $16.9k$ 3-multigrams and only $3.1k$ 2-multigrams. 

Finally, we analyze the system complexity according to the search graph size (number of states and transitions in the FST automaton $S$ defined in equation \ref{eq:fst}), the search graph volume on disk and the decoding time. Besides, the memory and the processing time during the decoding process depend on the search graph size (\cite{mohri2002weighted}). Note that complexities are computed from the unified FR+EN lexicon and that the character language model is a 9-gram model (in contrast to previous results based on a 10-gram model) to provide comparable complexities. 
\begin{table}[h!]
\centering
\caption{Size of decoding search graph (composed automaton S defined in equation \ref{eq:fst}) for  different language model with a small lexicon.}
\label{tbl:complexity-small}
\begin{tabular}{|c|c|c|c|}
\hline
\begin{tabular}[c]{@{}c@{}}Language models vs. \\ complexity indicators\end{tabular} & \begin{tabular}[c]{@{}c@{}}number of states\\ ($\times 10^6$)\end{tabular} & \begin{tabular}[c]{@{}c@{}}number of arcs\\ ($\times 10^6$)\end{tabular} & \begin{tabular}[c]{@{}c@{}}Volume on\\ disk (MB)\end{tabular} \\ \hline
Words & 1.37 & 3.46 & 71.9 \\ \hline
m5gram & 1.25 & 3.15 & 65.5 \\ \hline
m4gram & 1.20 & 3.04 & 63.1 \\ \hline
m3gram & 1.11 & 2.81 & 58.2 \\ \hline
m2gram & 0.93 & 2.40 & 49.5 \\ \hline
characters & 0.54 & 1.48 & 30.2 \\ \hline 
\end{tabular}%
\end{table}
As illustrated in Table \ref{tbl:complexity-small}, the number of states and transitions is highly reduced using multigram language models compared to the word-based language models. 
For instance, the 2-multigrams model reduces the sum of states and transitions of the search graph by $31\%$. The volume on the disk is also divided by $1.45$. 
\begin{table}[h!]
\centering
\caption{Size of decoding search graph (composed automaton S defined in equation \ref{eq:fst}) for  different language model with a large lexicon.}
\label{tbl:complexity-large}
\begin{tabular}{|c|c|c|c|}
\hline
\begin{tabular}[c]{@{}c@{}}Language models vs. \\ complexity indicators\end{tabular} & \begin{tabular}[c]{@{}c@{}}number of states\\ ($\times 10^6$)\end{tabular} & \begin{tabular}[c]{@{}c@{}}number of arcs\\ ($\times 10^6$)\end{tabular} & \begin{tabular}[c]{@{}c@{}}Volume on\\ disk (GB)\end{tabular} \\ \hline
Words & 178 & 447 & 9.3 \\ \hline
m5gram & 126 & 317 & 6.6 \\ \hline
m4gram & 108 & 269 & 5.6 \\ \hline
m3gram & 72 & 186 & 4.1 \\ \hline
m2gram & 65 & 166 & 3.4 \\ \hline
characters & 20 & 50 & 1 \\ \hline
\end{tabular}%
\end{table}
Similar comments can be made on a large lexicon with more significant results (Table \ref{tbl:complexity-large}). Dealing with 2-multigrams allows to reduce the sum of states and transitions of the search graph by $63\%$ compared to word model. The volume on disk is also reduced by a factor of $2.7$. Whatever the lexicon size, the character-based model is lighter than the other models but it never produces the best results in our experiments and it can perform sometimes poorly (as shown on the EN dataset in Figure \ref{fig:largeFREN}). 
\begin{table}[h!]
\centering
\caption{Decoding time on the RIMES validation set.}
\label{tbl:rtfactor}
\begin{tabular}{|c|c|c|}
\cline{2-3}
\hline
\begin{tabular}[c]{@{}c@{}}Language models \\ (LMs) \end{tabular} & \begin{tabular}[c]{@{}c@{}}Decoding time with \\ small LMs \\ (Minutes:Seconds) \end{tabular} & \begin{tabular}[c]{@{}c@{}}Decoding time with \\ large LMs \\ (Minutes:Seconds)  \end{tabular} \\ \hline
Words & 04:28 & 07:59  \\ \hline
m5gram & 03:39 & 07:32 \\ \hline
m4gram & 03:38 & 07:55 \\ \hline
m3gram & 03:34 & 07:14 \\ \hline
m2gram & 03:23 & 06:14  \\ \hline
characters & 03:14 & 08:18  \\ \hline
\end{tabular}%
\end{table}
Table \ref{tbl:rtfactor} shows the decoding time 
on the FR validation set, which relates to a measure computed on $1.7$ millions of frames. The decoding time is reduced using multigram language models compared to traditional language models based on words. For instance, using a 5-multigram language model reduces the time by $18\%$ and by $24\%$ using 2-multigrams on the small lexicon while there is a reduction of $22\%$ using 2-multigrams on the large lexicon. 

\begin{table}[h]
\caption{Performance comparison of the proposed system with results reported by other studies on the RIMES \& IAM test datasets.}
\centering
\begin{tabular}{|l|l|l|}
\hline
System & WER (\%) on RIMES & WER (\%) on IAM \\ \hline
Our unified system & 9.8 (2-multigrams)  & 11.2 (5-multigrams) \\ \hline
Our specialized system & 10.8 (2-multigrams) & 13.5 (5-multigrams) \\ \hline
\cite{voigtlaender2016handwriting} & 9.6 & 9.3  \\ \hline
\cite{bluche2015deep} & 11.8 & 11.9 \\ \hline
\cite{bluche2017gated} & 7.9 & 10.5 \\ \hline
\end{tabular}%
\label{tbl:CMP}
\end{table}

To conclude, we compare our multilingual recognition system with state-of-the art systems (Table \ref{tbl:CMP}). In our system, the optical model has a simpler architecture compared to state-of-the-art systems while these systems rest on language models based on words or hybrid models of words and characters. Multilingual systems based on multigrams reach performance closed to most of the state-of-the-art systems, while the system complexity is reduced. The specialized system presented in Table \ref{tbl:CMP} refer to a selective approach where the right specialized model is always selected (an ideal case). We show that, whatever the test set (FR or EN), our unified system makes profit of the combination of the languages to outperform the selective approach.



\section{Conclusion}

We presented an end-to-end unified multilingual system for handwriting recognition where both the optical model and the language model are trained on datasets composed of examples from several languages. Our proposal allows, on the one hand, to optimize a unique system, whatever the languages that are in training and, on the other hand, to avoid the use of a decision process for selecting one specialized system trained on a specific language. 

Our unified optical model is optimized to recognize a unified character set. In case of languages of the same origin, unifying the character set reduces the system complexity and increases the number of training examples per character classes that are shared between the languages. While traditional language models are based on words, which can become intractable in case of unified lexicons, we proposed to build a language model based on sub-lexical units, called multigrams. Dealing with multigrams has many advantages: the multigrams are obtained using a data-driven process without the need of linguistic expertise; it reduces the model complexity compared to words; finally, it allows a better modeling of long dependencies than with characters. 

Our experiments on English and French languages with small or large lexicons, highlighted that the optical model benefits from the language unification and provides significant improvements compared to specialized systems. A major contribution is to show that unifying languages that have some similarities does not affect the language models which provide similar results than the specialized language models. In addition, dealing with multigrams allows to improve the traditional language models based on words or characters. Finally, our system reaches state-of-the art performance with a unique and less complex system. 




\section*{Acknowledgement}

This work is founded by the French government, region Normandie and the European Union. Europe acts in Normandy with the European Regional Development Fund (ERDF).

\bibliographystyle{model2-names}
\bibliography{sample}



\end{document}